\documentclass[journal]{IEEEtran}
\usepackage{blindtext}
\usepackage{graphicx}
\usepackage{natbib} 
\usepackage[cmex10]{amsmath}
\usepackage{algorithmic}
\usepackage{array}
\usepackage{amssymb}
\usepackage{mdwmath}
\usepackage{mdwtab}
\usepackage[]{algorithm2e}
\usepackage{lmodern}
\usepackage{eqparbox}

\newcommand{\BigO}[1]{\ensuremath{\operatorname{O}\left(#1\right)}}
\newcommand{\nosemic}{\renewcommand{\@endalgocfline}{\relax}}
\newcommand{\dosemic}{\renewcommand{\@endalgocfline}{\algocf@endline}}

\begin{document}
\title{Introduction to Gravitational Clustering}

\author{Armen Aghajanyan}
\markboth{Pattern Analysis and Machine Intelligence,~Vol.~X, No.~Y, February~2015}%
{Shell \MakeLowercase{\textit{et al.}}: Bare Demo of IEEEtran.cls for Journals}

\maketitle

\begin{abstract}
The downfall of many supervised learning algorithms, such as neural networks, is the inherent need for a large amount of training data \citep{nndown}. Although there is a lot of buzz about big data, there is still the problem of doing classification from a small data-set. Other methods such as support vector machines, although capable of dealing with few samples, are inherently binary classifiers \citep{svmbinary}, and are in need of learning strategies such as One vs All in the case of multi-classification. In the presence of a large number of classes this can become problematic. In this paper we present, a novel approach to supervised learning through the method of clustering. Unlike traditional methods such as K-Means \citep{kmeans}, Gravitational Clustering does not require the initial number of clusters, and automatically builds the  clusters, individual samples can be arbitrarily weighted and it requires only few samples while staying resilient to over-fitting. 
\end{abstract}


\section{Introduction} 
The name of this algorithm is derived from the metaphor that the algorithm was built upon. Each cluster is symbolic of a planet,and each planet has a mass and a radius as well as the class that it represents. But unlike real life planets, our planets are static with respect to other planets. The process of training can be conceptually thought of as building a universe. The process of predicting is simply placing a mass in the universe and tracing what planet it will appear on. 
\\
\\
This algorithm exhibits three nice properties: 
\begin{enumerate}
\item Ability to learn from a few samples.
\item Ability to weight the importance of training vectors.
\item The nature of the algorithm makes it resilient to overfitting.
\end{enumerate}
The ability to weight the importance of training vectors as well as the ability to learn from a few samples allows us to model a system that supports the notion of prototypes, e.g. Eleanor Rosch \citep{proto}(P. 41).
\section{Definition}
Let us start by mathematically defining what each one of our symbolic structures will be. The most important structure is our cluster or our planet. We will define the planet as containing a dynamic mass $m$, dynamic radius $r$, dynamic position $\overrightarrow{x}$ and a static class $\theta$. Mathematically:
\begin{equation}
	\begin{array}
		{lcl} 
		m & \in & \mathbb{R} \\ 
		r & \in & \mathbb{R} \\ 
		\overrightarrow{x} & \in & \mathbb{R}^n \\ 
		\theta & \in & \mathbb{Z} \\ \\
		 \mathbb{P} & = & \{m,r,\overrightarrow{x},\theta\}
	\end{array}
\end{equation}
Our universe will simply consist of a set of planets. The universe will also hold a couple of global constants. The initial radius of a planet that has just been created which we will denote with $r'$. The so called percent step, which represents the amount a test mass moves before recalculating the new forces on the test mass. We will denote this with the Greek $\alpha$. The amount of steps taken or iterations will be denoted with $\beta$. The distance between planets will be calculated with the function denoted $D(\overrightarrow{x},\overrightarrow{y})$.  
\section{Training Model}
One of the better aspects of the model is its ability to rate your feature vectors. To do so, let us define a hybrid feature vector $h$.
\begin{equation}
h=\{\overrightarrow{x}, m,\theta\}
\end{equation} 
The $m$ variable allows us to rate the value of the feature vector. For example if you have a probabilistic diagnosis, each feature vector will contain the class of the diagnosis as well as the probability of the diagnosis represented by the mass. The training is quite simple. 
Below is the pseudo-code.
\begin{algorithm}
 nearplanets $\leftarrow$ Find Planets in Radius of $\overrightarrow{h}_x$\;
 nearplanets $\leftarrow$ nearplanets where $P_\theta = h_\theta$ \;
 \eIf{nearplanets is Empty}
 {
	Universe Add Planet $\{m = h_m,r = r',\overrightarrow{x} = \overrightarrow{h}_x, \theta = h_\theta\}$ 
 }
 {
 p $\leftarrow$ planet that generates most force $\in$ nearplanets \;
 Universe update p $\leftarrow$  $\{m = p_m + h_m,r = m\frac{p_r}{p_m}, \overrightarrow{x} = \frac{p_m}{m}\overrightarrow{p}_x+ \frac{h_m}{m}\overrightarrow{h}_x \}$ 
 }
\caption{Training Algorithm}
\end{algorithm}
\\
The new position is a weighted sum of the two position vectors with respect to their weight.
\subsection{Asymptotic Analysis}
Our training simply traverses through all of the planets in the universe and computes the distance from the training sample. Saying $N$ is the amount of planets and $D$ is the dimensionality of our feature vectors. Assuming that the planet exists, we get
\begin{equation}
\BigO{D*N} = \BigO{N}
\end{equation}
Using a KD-Tree 
\citep{kdtree} will allow us to train with the average asymptotic of 
\begin{equation}
\BigO{D*\log{N}} = \BigO{\log{N}} 
\end{equation}
On the flip side, assuming we have to add the planet:
\begin{equation}
\BigO{D*N + N_{near}} = \BigO{N + N_{near}}
\end{equation}
KD-Tree \citep{kdtree}
\begin{equation}
\BigO{D*\log{N} + N_{near}} = \BigO{\log{N} + N_{near}} 
\end{equation}
This is the asymptotic of adding a single train vector. Stating that $N_s$ is the number of samples we end up with the final equation being.
\begin{equation}
\BigO{N_s(\log{N} + N_{n})}
\end{equation}

\subsection{Comparison of Training Times}
\begin{table}[h]
\begin{tabular}{ |  l  |l|l|l|l|}
\hline
                                                               & \begin{tabular}[c]{@{}l@{}}Gravitational\\ Clustering\end{tabular} & K-Means              & SVM                        & Decision Trees                                \\ \hline
Big O                                                          & 
 {\fontsize{4}{12} \selectfont $\BigO{N_s(\log{N} + N_{n})}$} 

                                       & {\fontsize{4}{12} \selectfont 
 
 $O(n^{Dk+1}\log{n})$}& {\fontsize{4}{12} \selectfont  $O(n^{3})$ }& {\fontsize{4}{12} \selectfont $O(n_{s}D\log(n_{s}))$ } \\ \hline
\begin{tabular}[c]{@{}l@{}}Online \\ Training\end{tabular}     & Yes                                                                & Yes                  & No                         & Partial                                       \\ \hline
\begin{tabular}[c]{@{}l@{}}Variant \\ Importance\end{tabular} & Yes                                                                & No                   & No                         & No                                            \\ \hline
\end{tabular}
\end{table}

\begin{itemize}
	\item $N_n$ is synonymous with $N_{near}$
	\item $N_s$ is synonymous with $N_{samples}$
\end{itemize}
\section{Simulation Testing Model}
Metaphorically, predicting the class of a new point is equivalent to dropping a piece of mass into the universe and tracing the mass until it collides with a planet. In this metaphor, we assume that the planets are infinitely small and therefore there will be no interference. Our test point will simply be defined as $l = \{\overrightarrow{x}\}$ Let us first define getting the normalized directional force vector. Recall from physics that the gravitational force between two planets is
\begin{equation}
F = \mathbb{G}\frac{m_1m_2}{r^2}
\end{equation}
In our case, we will assume that the mass of each test point is equal 
to every other, therefore we can disregard the mass. We can also remove the $\mathbb{G}$ constant. Our hybrid force equation per planet $p$ is now:
\begin{equation}
F = \frac{p_m}{r^2}
\end{equation}
Where r is $D(\overrightarrow{p}_x,\overrightarrow{l}_x)$. We define the total normalized force on our test mass with the custom equation. 
\begin{equation}
\begin{array}
		{lcl} 
		 F_{net} = \sum_{p \in Universe}\frac{p_m*(\overrightarrow{p}_x -\overrightarrow{l}_x)}{r^2} \\
		 F_{norm} = \frac{\alpha}{||F_{net}||} F_{net}
	\end{array}
\end{equation}
To restate, $\alpha$ is the percent step taken with respect to the force. 
Now let us describe the simulation algorithm:

\begin{algorithm}
  pos $\leftarrow$ $\overrightarrow{l}_x$\;
  \For{i in i $[0,\beta]$ step 1}
  {
	force $\leftarrow$ $\sum_{p \in Universe}\frac{p_m*(\overrightarrow{p}_x - pos)}{r^2}$ \;
	\;
	 norm $\leftarrow$ $\frac{\alpha}{||force||} force$ \;
	 pos $\leftarrow$ pos + norm
  }
  nearplanets $\leftarrow$ Find Planets in Radius of pos\;
  \eIf{nearplanets is not Empty}
 {
 	return $\mathbf{mode}$[nearplanets $\theta$]
 }
 {
 return [planet closest to pos] $\theta$
 }
\end{algorithm}
\subsection{Asymptotic Analysis of Simulation Testing Model}
Let us state that $N$ is the number of planets and $D$ is the dimensionality of our feature vector. Calculating the force takes up 
\begin{equation}
\BigO{4D*N} = \BigO{N} 
\end{equation}
The 4 comes from the vector arithmetic that needed to be done. One subtraction, one multiplication, one distance squared, one division. The N term came from the summation. The total simulation next becomes.
\begin{equation}
\BigO{7D*N *\beta + N} 
\end{equation}
The 3 more D terms come from: finding the magnitude, multiplying by force (simultaneously multiplying by $\alpha$) and the update summation. The next N came from finding the planets with the radius containing pos. We can disregard the final if statement since they do not directly affect N. We get:
\begin{equation}
\BigO{7D*N *\beta + N} = \BigO{N(7D*\beta+1) }=\BigO{N}  
\end{equation} 

\section{Probabilistic Non-Simulating Model}
We propose an different method of computing the class of the test point, without the need of simulation and through purely statistical methods. We first make an assumption that a planet or cluster is normally distributed from the center and the standard deviation is some function of the radius of the planet $\sigma(p_r)$. Therefore let us the define the probability density function.

\begin{equation}
\mathbf{PDF_p} = \frac{1}{2\pi*\sigma(p_r)} e ^ \frac{-D(\overrightarrow{p}_x,\overrightarrow{l}_x)^2}{2\sigma(p_r)^2}
\end{equation}

Now to define our prediction equation:
\begin{equation}
\mathbf{MAX_\theta[} \prod_{p \ | \ p_\theta = \theta_n}^{Universe} \frac{1}{2\pi*\sigma(p_r)} e ^ \frac{-D(\overrightarrow{p}_x,\overrightarrow{l}_x)^2}{2\sigma(p_r)^2} \mathbf{]}
\end{equation}
To account for the fact that different classes have different amounts of planets, we will transform this function into:
\begin{equation}
\mathbf{MAX_\theta[} \frac{ \log {\prod_{p \ | \ p_\theta = \theta_n}^{Universe}   e ^ {(\frac{-D(\overrightarrow{p}_x,\overrightarrow{l}_x)^2}{p_m2\sigma(p_r)^2})} \mathbf{]}}}{|p \ | \ p_\theta = \theta_n|}
\end{equation}
We removed the normalization constant, due to the fact that this is a relative measure. The bottom of the fraction is the number of planets per class which insures that there is no bias due to the different amounts of clusters with varying radius's. The mass term is added to insure that greater planets have a greater impact on the rating. \\ \\
Through trial and error we found the best function for  $\sigma(p_r)$ was simply  $p_r^2$. 
\\ \\ 
The asymptotic will simply be 
\begin{equation}
\BigO{DN} = \BigO{N}
\end{equation}
\section{Testing Results}
We tested the algorithm out on the Wisconsin breast cancer data-set \citep{breastcancer} \citep{wisconsin}. Below are the results.
\begin{table}[h]
\tabcolsep=6mm
\begin{tabular}{|l|l|l|}
\hline
Gravitational Clustering & \begin{tabular}[c]{@{}l@{}}$r' = 50$\\ $\alpha = 0.01$\\ $\beta = 100$\end{tabular} & \begin{tabular}[c]{@{}l@{}}$r' = 5000$\\ $\alpha = 0.001$\\  $\beta = 1000$\end{tabular} \\ \hline
Simulated Model          & 89.65\%                                                                         & 90.59\%                                                                                  \\ \hline
Probabilistic Model      & 92.78\%                                                                         & 72.41\%                                                                                  \\ \hline
\end{tabular}
\end{table}

It is interesting to note that the larger the clusters and smaller the amount of clusters the less accurate the probabilistic model will be. Unless of course the clusters perfectly model the data that they encapsulate.
\\ 
We continued our testing by comparing the outputs of some popular out of the box methods. All the other algorithms were implemented in the scikit-learn library \citep{sklearn}. The data-sets we used were the popular Iris data-set \citep{wisconsin}, digits data-set\citep{sklearn}, Ollivetti data-set \citep{olivetti}.

\begin{table}[h]
\begin{tabular}{|l|l|l|l|l|l|}
\hline
 &  & \begin{tabular}[c]{@{}l@{}}Algorithm \\ \end{tabular} &  &  &  \\ \hline
\begin{tabular}[c]{@{}l@{}}Data-sets\end{tabular} & GC Prob & GC Sim & \begin{tabular}[c]{@{}l@{}}SVM\\ (poly)\end{tabular} & \begin{tabular}[c]{@{}l@{}}SVM\\ (rbf)\end{tabular} & \begin{tabular}[c]{@{}l@{}}Naive Bayes\\ (Gaussian)\end{tabular} \\ \hline
Iris & 98.41\% & 96.82\% & 94.66\% & 97.33\% & 96\% \\ \hline
Digits & 86.95\% & 91.04\% & 98.99\% & 25.61\% & 83.85\% \\ \hline
Olivetti & 65.5\% & 77.5\% & 7.5\% & 8.5\% & 99.5\% \\ \hline
\end{tabular}
\end{table}
To show that our algorithm can handle very few samples, we tested the following data-sets again, but this time we only used 1 sample per each class as the training data. Below are the results.
\begin{table}[h]
\begin{tabular}{|l|l|l|}
\hline
 & Accuracy Per Data-set &  \\ \hline
\begin{tabular}[c]{@{}l@{}}Algorithm\\ Type\end{tabular} & GC Prob & GC Sim \\ \hline
Iris & 93.33\% & 92.00\% \\ \hline
Digits & 59.96\% & 58.18\% \\ \hline
Olivetti & 63.5\% & 53.75\% \\ \hline
\end{tabular}
\end{table}
\section{Conclusion}
In this paper we introduced a novel technique to clustering and supervised learning that can learn from a few samples, while maintaining a low asymptotic run-time and inherently allowing for arbitrary sample weighting. We compared it to current techniques for classification and showed both the strengths of the algorithm as well as the weaknesses. From the test results we can infer that our algorithm acts consistently in both low and high dimensional data, as well as staying consistent in a range of multi-class data-sets. All the code written, including the tests and the algorithm itself can be found on https://github.com/ArmenAg/GravitationalClustering/

\begin{center}
\textbf{Thank you for reading.}
\end{center}

\bibliography{main}

\end{document}